\documentclass[10pt,twocolumn,letterpaper]{article}

\usepackage{iccv}
\usepackage{times}
\usepackage{epsfig}
\usepackage{graphicx}
\usepackage{amsmath}
\usepackage{amssymb}

\usepackage{caption}
\usepackage{subcaption}
\usepackage{booktabs}

\usepackage[pagebackref=true,breaklinks=true,letterpaper=true,colorlinks,bookmarks=false]{hyperref}

\newtheorem{theorem}{Theorem}
\iccvfinalcopy %

\def\iccvPaperID{5819} %

\ificcvfinal\fi

\begin{document}

\title{Efficient Diffusion Training via Min-SNR Weighting Strategy}

\author{Tiankai Hang$^1$,
Shuyang Gu$^2$\thanks{ Corresponding authors.},
Chen Li$^3$,
Jianmin Bao$^2$,
Dong Chen$^2$,\\
Han Hu$^2$,
Xin Geng$^1$,
Baining Guo$^1$\footnotemark[1]\\
$^1$Southeast University, $^2$Microsoft Research Asia, \\ $^3$National Key Laboratory of Human-Machine Hybrid Augmented Intelligence,\\National Engineering Research Center for Visual Information and Applications,\\ and Institute of Artificial Intelligence and Robotics, Xi'an Jiaotong University \\
{\tt\small \{tkhang,xgeng,307000167\}@seu.edu.cn,\{shuyanggu,t-chenli1,jianmin.bao,doch,hanhu\}@microsoft.com}
}

\maketitle
\ificcvfinal\thispagestyle{empty}\fi

\begin{abstract}

Denoising diffusion models have been a mainstream approach for image generation, however, training these models often suffers from slow convergence. In this paper, we discovered that the slow convergence is partly due to conflicting optimization directions between timesteps. To address this issue, we treat the diffusion training as a multi-task learning problem, and introduce a simple yet effective approach referred to as Min-SNR-$\gamma$. This method adapts loss weights of timesteps based on clamped signal-to-noise ratios, which effectively balances the conflicts among timesteps. Our results demonstrate a significant improvement in converging speed, 3.4$\times$ faster than previous weighting strategies. It is also more effective, achieving a new record FID score of 2.06 on the ImageNet $256\times256$ benchmark using smaller architectures than that employed in previous state-of-the-art. The code is available at \href{https://github.com/TiankaiHang/Min-SNR-Diffusion-Training}{https://github.com/TiankaiHang/Min-SNR-Diffusion-Training}.

\end{abstract}

\section{Introduction}

In recent years, denoising diffusion models~\cite{sohl2015nonequ,ho2020ddpm,2021Scoresde,nichol2021iddpm} have emerged as a promising new class of deep generative models due to their remarkable ability to model complicated distributions. Compared to prior Generative Adversarial Networks (GANs), diffusion models have demonstrated superior performance across a range of generation tasks in various modalities, including text-to-image generation~\cite{ramesh2022dalle2,saharia2022imagen,rombach2022ldm,gu2022vqdiffusion}, image manipulation~\cite{kim2021diffusionclip,meng2021sdedit,brooks2022instructpix2pix,yang2022paint}, video synthesis~\cite{Ho2022imagenvideo,Singer2022makeavideo,ho2022video}, text generation~\cite{li2022diffusion-lm,gongsequence,zhu2022exploring}, 3D avatar synthesis~\cite{poole2022dreamfusion,wang2022rodin}, etc.
A key limitation of present denoising diffusion models is their slow convergence rate, requiring substantial amounts of GPU hours for training~\cite{rombach2022ldm,ramesh2022hierarchical}. This constitutes a considerable challenge for researchers seeking to effectively experiment with these models.

 In this paper, we first conducted a thorough examination of this issue, revealing that the slow convergence rate likely arises from conflicting optimization directions for different timesteps during training. In fact, we find that by dedicatedly optimizing the denoising function for a specific noise level can even harm the reconstruction performance for other noise levels, as shown in Figure~\ref{fig:ft-loss-curve}. This indicates that the optimal weight gradients for different noise levels are in conflict with one another. Given that current denoising diffusion models~\cite{ho2020ddpm,dhariwal2021adm,nichol2021iddpm,rombach2022ldm} employ shared model weights for various noise levels, the conflicting weight gradients will impede the overall convergence rate, if without careful consideration on the balance of these noise timesteps.

\begin{figure}[t]
\begin{center}
\vspace{-0.1cm}
    \includegraphics[width=0.47\textwidth]{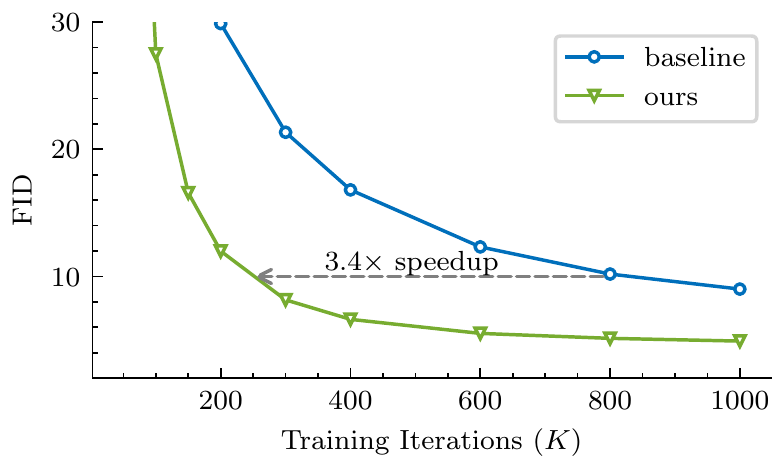}
\end{center}
\vspace{-0.7cm}
   \caption{By leveraging a non-conflicting weighting strategy, our method can converge 3.4 times faster than baseline, resulting in superior performance.}
\label{fig:teaser}
\end{figure}

To tackle this problem, we propose the \textbf{Min-SNR-$\gamma$} loss weighting strategy. This strategy treats the denoising process of each timestep as an individual task, thus diffusion training can be considered as a multi-task learning problem. To balance various tasks, we assign loss weights for each task according to their difficulty. Specifically, we adopt a clamped signal-to-noise ratio (SNR) as loss weight to alleviate the conflicting gradients issue. By organizing various timesteps using this new weighting strategy, the diffusion training process can converge much faster than previous approaches, as illustrated in Figure~\ref{fig:teaser}.

Generic multi-task learning methods usually seek to mitigate conflicts between tasks by adjusting the loss weight of each task based on their gradients. One classical approach~\cite{desideri2012mgda,sener2018mto}, Pareto optimization, aims to seek a gradient descent direction to improve all the tasks. However, these approaches differ from our Min-SNR-$\gamma$ weighting strategy in three aspects: 1) \textbf{Sparsity}. Most previous studies in the generic multi-task learning field have focused on scenarios with a small number of tasks, which differs from the diffusion training where the number of tasks can be up to thousands. As in our experiments, Pareto optimal solutions in diffusion training tend to set loss weights of most timesteps as 0. In this way, many timesteps will be left without any learning, and thus harm the entire denoising process. 2) \textbf{Instability}. The gradients computed for each timestep in each iteration are often noisy, owing to a limited number of samples for each timestep. This hampers the accurate computation of Pareto optimal solutions. 3) \textbf{Inefficiency}. The calculation of Pareto optimal solutions is time-consuming, significantly slowing down the overall training.

Our proposed Min-SNR-$\gamma$ strategy is a predefined global step-wise loss weighting setting, instead of run-time adaptive loss weights for each iteration as in the original Pareto optimization, thus avoiding the sparsity issue. Moreover, the global loss weighting strategy eliminates the need for noisy computation of gradients and the time-consuming Pareto optimization process, making it more efficient and stable. Though suboptimal, the global strategy can be also almost as effective: Firstly, the optimization dynamics of each denoising task are largely shaped by the task's noise level, without the need to account for individual samples too much. Secondly, after a moderate number of iterations, the gradients of the majority subsequent training process become more stable, thus it can be approximated by a stationery weighting strategy.

To validate the effectiveness of the Min-SNR-$\gamma$ weighting strategy, we first compute its Pareto objective value and compare it with the optimal step-wise loss weights obtained by directly solving the Pareto problem. Together, we also compare it with several conventional loss weighting strategies, including constant weighting, SNR weighting, and SNR with an lower bound. Figure~\ref{fig:object-value} shows that our Min-SNR-$\gamma$ weighting strategy produces Pareto objective values almost as low as the optimal one, significantly better than other existing works, indicating a significant alleviation of the gradient conflicting issue. As a result, the proposed weighting strategy not only converges much faster than previous approaches, but is also effective and general for various generation scenarios. It achieves a new record of FID score 2.06 on the ImageNet 256$\times$256 benchmark, and proves to also improve models using other prediction targets and network architectures.

Our contributions are summarized as follows:
\begin{itemize}
    \item We have uncovered a compelling explanation for the slow convergence issue in diffusion training: a conflict in gradients across various timesteps.
    \item %
    We have proposed a new loss weighting strategy for diffusion model training, which greatly mitigates the conflicting gradients across timesteps and results in a marked acceleration of convergence speed.
    \item We have established a new FID score record on the ImageNet $256\times 256$ image generation benchmark.
\end{itemize}

\section{Related Works}
\noindent\textbf{Denoising Diffusion Models.}
 Diffusion models~\cite{ho2020ddpm, song2019generative,dhariwal2021adm} are strong generative models, particularly in the field of image generation, due to their ability to model complex distributions. This advantage has led to superiority over previous GAN models in terms of both high-fidelity and diversity of generated images~\cite{dhariwal2021adm, karras2022edm, Nichol2021glide, ramesh2022dalle2, rombach2022ldm, saharia2022imagen}. Besides, diffusion models also show great success in text-to-video generation~\cite{Ho2022imagenvideo,Singer2022makeavideo,villegas2022phenaki}, 3D Avatar generation~\cite{poole2022dreamfusion,wang2022rodin}, image to image translation~\cite{parmar2023zeroI2I}, image manipulation~\cite{brooks2022instructpix2pix,kim2021diffusionclip}, music generation~\cite{huang2023noise2music}, and even drug discovery~\cite{xu2022geodiff}. The most widely used network structure for diffusion models in the field of image generation is UNet~\cite{ho2020ddpm, dhariwal2021adm, Nichol2021glide, nichol2021iddpm}. Recently, researchers have also explored the use of Vision Transformers~\cite{vit} as an alternative, with U-ViT~\cite{bao2022uvit} borrowing the skip connection design from UNet~\cite{unet} and DiT~\cite{peebles2022dit} leveraging Adaptive LayerNorm and discovering that the zero initialization strategy is critical for achieving state-of-the-art class-conditional ImageNet generation results.

\noindent\textbf{Improved Diffusion Models.} 
Recent studies have tried to improve the diffusion models from different perspectives. Some works aim to improve the quality of generated images by guiding the sampling process~\cite{dhariwal2021diffusion,ho2022classifier}. Other studies propose fast sampling methods that require only a dozen steps~\cite{ddim, Liu2022PNMD, Lu2022DPMSolverAF, karras2022edm} to generating high-quality images. Some works have further distilled the diffusion models for even fewer steps in the sampling process~\cite{salimans2022distillprogressive,meng2022ondistillation}. Meanwhile, some researchers~\cite{ho2020ddpm,karras2022edm,chen2023importance} have noticed that the noise schedule is important for diffusion models. Other works~\cite{nichol2021iddpm, salimans2022distillprogressive} have found that different predicting targets from denoising networks affect the training stability and final performance. Finally, some works~\cite{feng2022ernie2,balaji2022eDiff-I} have proposed using the Mixture of Experts (MoE) approach to handle noise from different levels, which can boost the performance of diffusion models, but require a larger number of parameters and longer training time.

\noindent\textbf{Multi-task Learning.} 
The goal of Multi-task learning (MTL) is to learn multiple related tasks jointly so that the knowledge contained in a task can be leveraged by other tasks. One of the main challenges in MTL is negative transfer~\cite{crawshaw2020multi}, means the joint training of tasks hurts learning instead of helping it. From an optimization perspective, it manifests as the presence of conflicting task gradients. To address this issue, some previous works~\cite{yu2020pcgrad,wang2020gradvac,chen2020graddrop} try to modulate the gradient to prevent conflicts. Meanwhile, other works attempt to balance different tasks through carefully design the loss weights~\cite{chen2018gradnorm,kendall2018multi}. GradNorm~\cite{chen2018gradnorm} considers loss weight as learnable parameters and updates them through gradient descent. Another approach MTO~\cite{desideri2012mgda,sener2018mto} regards the multi-task learning problem as a multi-objective optimization problem and obtains the loss weights by solving a quadratic programming problem.

\section{Method}

\subsection{Preliminary}\label{sec:preliminary}

Diffusion models consist of two processes: a forward noising process and a reverse denoising process. 
We denote the distribution of training data as $p(\mathbf{x}_0)$.
The forward process is a Gaussian transition, gradually adds noise with different scales to a real data point $\mathbf{x}_0\sim p(\mathbf{x}_0)$ to obtain a series of noisy latent variables $\{ \mathbf{x}_1,\mathbf{x}_2, \ldots, \mathbf{x}_T \}$:
\begin{align}
    q(\mathbf{x}_t | \mathbf{x}_0) &= \mathcal{N}( \mathbf{x}_t; \alpha_t \mathbf{x}_0, \sigma_t^2 \mathbf{I}) \\
    \mathbf{x}_t &= \alpha_t \mathbf{x}_0 + \sigma_t \boldsymbol{\epsilon} \label{eq:diffusion}
\end{align}
where $\boldsymbol{\epsilon}$ is the noise sampled from Gaussian distribution $\mathcal{N}(0, \mathbf{I})$.
The noise schedule $\sigma_t$ denotes the magnitude of noise added to the clean data at $t$ timestep. It increases monotonically with $t$. In this paper, we adopt the standard variance-preserving diffusion process, where $\alpha_t = \sqrt{1 - \sigma_t^2}$.

The reverse process is parameterized by another Gaussian transition, gradually denoises the latent variables and restores the real data $\mathbf{x}_0$ from a Gaussian noise:
\begin{equation}
    p_\theta(\mathbf{x}_{t-1} | \mathbf{x}_t) = \mathcal{N}(\mathbf{x}_{t-1}; \mathbf{\hat{\mu}}_\theta (\mathbf{x}_{t}), \hat{\Sigma}_\theta(\mathbf{x}_{t})).
\end{equation}
$\mathbf{\hat{\mu}}_\theta$ and $\hat{\Sigma}_\theta$ are predicted statistics. 
Ho et al.~\cite{ho2020ddpm} set $\hat{\Sigma}_\theta(\mathbf{x}_{t})$ to the constant $\sigma_t^2\mathbf{I}$, and $\hat{\mu}_\theta$ can be decomposed into the linear combination of $\mathbf{x}_t$ and a noise approximation model $\hat{\epsilon}_\theta$. They find using a network to predict noise $\mathbf{\epsilon}$ works well, especially when combined with a simple re-weighted loss function:
\begin{equation}\label{equ:simple}
    \mathcal{L}_{\text{simple}}^t (\theta) = \mathbb{E}_{\mathbf{x}_0,\mathbf{\epsilon}}\left[
    \lVert \mathbf{\epsilon} - \hat{\epsilon}_\theta (\alpha_t \mathbf{x}_0 + \sigma_t \epsilon) \rVert_2^2
    \right].
\end{equation}
Most previous works~\cite{nichol2021iddpm,dhariwal2021adm,Nichol2021glide} follow this strategy and predict the noise. Later works~\cite{gu2022vqdiffusion,salimans2022distillprogressive} use another re-parameterization that predicts the noiseless state $x_0$:
\begin{equation}\label{equ:simple-x0}
    \mathcal{L}_{\text{simple}}^t (\theta) = \mathbb{E}_{\mathbf{x}_0,\mathbf{\epsilon}}\left[
     \lVert \mathbf{x}_0 - \hat{\mathbf{x}}_\theta (\alpha_t \mathbf{x}_0 + \sigma_t \epsilon) \rVert _2^2
    \right].
\end{equation}
And some other works~\cite{salimans2022distillprogressive,rombach2022ldm} even employ the network to directly predict velocity $v$. Despite their prediction targets being different, we can derive that they are mathematically equivalent by modifying their loss weights.

\subsection{Diffusion Training as Multi-Task Learning}
To reduce the number of parameters, previous studies~\cite{ho2020ddpm,nichol2021iddpm,dhariwal2021adm} often share the parameters of the denoising models across all steps.
However, it's important to keep in mind that different steps may have vastly different requirements. At each step of a diffusion model, the strength of the denoising varies. 
For example, easier denoising tasks (when $t\to 0$) may require simple reconstructions of the input in order to achieve lower denoising loss. This strategy, unfortunately, does not work as well for noisier tasks (when $t\to T$). 
Thus, it's extremely important to analyze the correlation between different timesteps.

\begin{figure}
    \centering
    \includegraphics[width=0.48\textwidth]{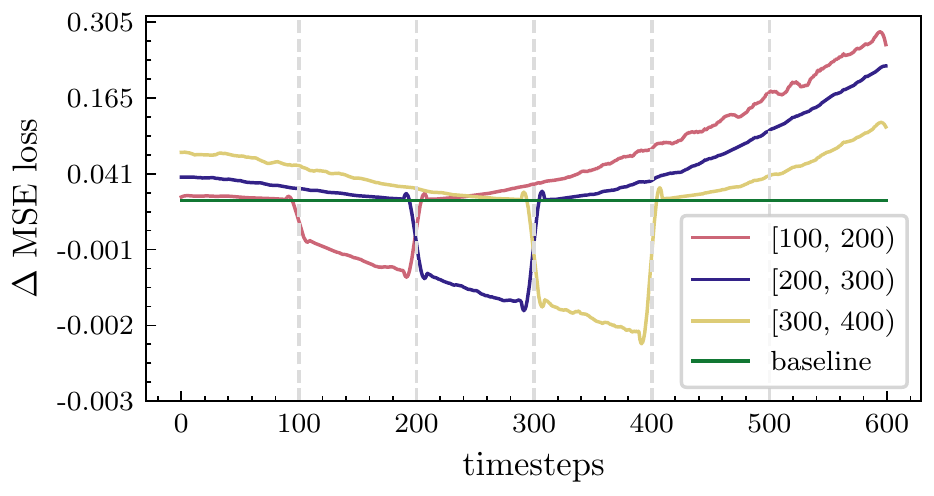}
    \vspace{-0.4cm}
    \caption{
    We finetune the diffusion model in specific ranges of timesteps:[100, 200), [200, 300), and [300, 400), then we investigate how it affects the loss in different timesteps. The surrounding timesteps may derive benefit from it, while others may experience adverse effects.
    }
    \label{fig:ft-loss-curve}
\end{figure}

In this regard, we conduct a simple experiment. We begin by clustering the denoising process into several separate bins. Then we finetune the diffusion model by sampling timesteps in each bin. Lastly, we evaluate its effectiveness by looking at how it impacted the loss of other bins. As shown in Figure~\ref{fig:ft-loss-curve}, we can observe that finetuning specific steps benefited those surrounding steps. However, it's often detrimental for other steps that are far away. This inspires us to consider whether \emph{we can find a more efficient solution that benefits all timesteps simultaneously}.

We re-organized our goal from the perspective of multitask learning. The training process of denoising diffusion models contains $T$ different tasks, each task represents an individual timestep. We denote the model parameters as $\theta$ and the corresponding training loss is $\mathcal{L}^t(\theta), t\in \{ 1,2,\ldots,T\}$. Our goal is to find a update direction $\delta \neq 0$, that satisfies:
\begin{equation}
    \mathcal{L}^t(\theta+\delta) \leq \mathcal{L}^t(\theta), \forall t \in \{1,2,\ldots,T\}.
\end{equation}
We consider the first-order Taylor expansion:
\begin{equation}
\mathcal{L}^t(\theta+\delta) \approx \mathcal{L}^t(\theta) + \left \langle\delta, \nabla_{\theta}\mathcal{L}^t(\theta)\right \rangle.
\end{equation}
Thus, the ideal update direction is equivalent to satisfy:
\begin{equation}
    \left \langle\delta, \nabla_{\theta}\mathcal{L}^t(\theta)\right \rangle \leq 0, \forall t \in \{1,2,\ldots,T\}.
    \label{eqn:optimization_objective}
\end{equation}

\subsection{Pareto optimality of diffusion models}

\begin{theorem}
Consider a update direction $\delta^*$:
\begin{equation}
\delta^* = -\sum_{t=1}^T w_t \nabla_{\theta} \mathcal{L}^t(\theta),
\end{equation}
of which $w_t$ is the solution to the optimization problem:
\begin{equation}
    \label{eq:opt}
    \min_{w^t} \left\{   \lVert \sum_{t=1}^T w^t \nabla_{\theta} \mathcal{L}^t(\theta)\rVert^2  | \sum_{t=1}^T w^t=1, w^t \geq 0\right\}
\end{equation}
If the optimal solution to the Equation~\ref{eqn:optimization_objective} exists, then $\delta^*$ should satisfy it. Otherwise, it means that we must sacrifice a certain task in exchange for the loss decrease of other tasks. In other words, we have reached the Pareto Stationary and the training has converged.
\end{theorem}

A more general form of this theorem was first proposed in~\cite{desideri2012mgda} and we leave a succinct proof in the appendix. 
Since diffusion models are required to go through all the timesteps when generating images. So any timestep should not be ignored during training. Consequently, a regularization term is included to prevent the loss weights from becoming excessively small. The optimization goal in Equation~\ref{eq:opt} becomes:
\begin{equation}
    \label{eq:reg-opt}
    \min_{w_t} \left \{ \lVert \sum_{t=1}^T w_t \nabla_{\theta} \mathcal{L}^t(\theta)\rVert_2^2 + \lambda \sum_{t=1}^T \lVert w_t \rVert_2^2 \right \}
\end{equation}
where $\lambda$ controls the regularization strength. %

To solve Equation~\ref{eq:reg-opt},~\cite{sener2018mto} leverages the Frank-Wolfe~\cite{frank1956frank} algorithm to obtain the weight $\{ w_t \}$ through iterative optimization. Another approach is to adopt Unconstrained Gradient Descent(UGD). Specifically, we re-parameterize $w_t$ through $\beta_t$:
\begin{equation}
    w_t = \frac{e^{\beta_t}}{Z}, Z=\sum_{t} e^{\beta_t}, \beta_t \in \mathbb{R}.
\end{equation}
Combined with Equation~\ref{eq:reg-opt}, we can use gradient descent to optimize each term independently:
\begin{equation}
    \label{eq:ugd-opt}
    \min_{\beta_t} \frac{1}{Z^2} \lVert \sum_{t=1}^T e^{\beta_t} \nabla_{\theta} \mathcal{L}_t(\theta)\rVert_2^2 + \frac{\lambda}{Z^2} \sum_{t=1}^T \lVert e^{\beta_t} \rVert_2^2
\end{equation}

\begin{figure}[t]
    \centering
    \includegraphics[width=0.45\textwidth]{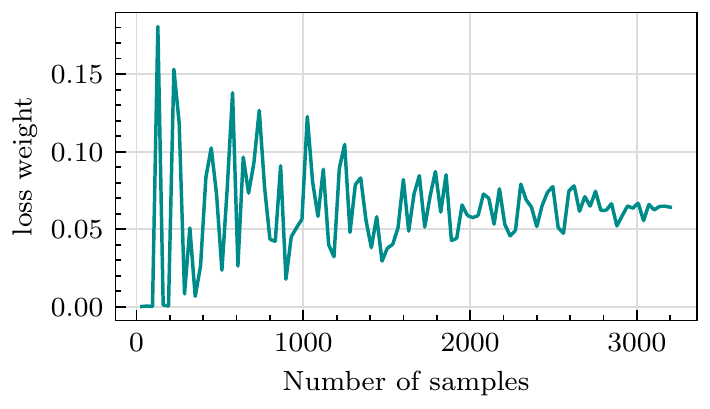}
    \vspace{-0.3cm}
    \caption{Demonstration of the instability of optimization-based weighting strategy. As the number of samples increases, the loss weight becomes stable, while the computation cost increases.
    }
    \vspace{-0.3cm}
    \label{fig:viz-alpha-instablity}
\end{figure}

However, whether leveraging the Frank-Wolfe or the UGD algorithm, there are two disadvantages:
1) \textit{Inefficiency}. Both of these two methods need additional optimization at each training iteration, it greatly increases the training cost.
2) \textit{Instability}. In practice, by using a limited number of samples to calculate the gradient term $\nabla_{\theta} \mathcal{L}^t(\theta)$, the optimization results are unstable(as shown in Figure~\ref{fig:viz-alpha-instablity}). In other words, the loss weights for each denoising task vary greatly during training, making the entire diffusion training inefficient.

\subsection{Min-SNR-$\gamma$ Loss Weight Strategy}\label{sec:different-loss-strategy}
In order to avoid the inefficiency and instability caused by the iterative optimization in each iteration, one possible attempt is to adopt a stationery loss weight strategy. 

To simplify the discussion, we assume that the network is reparametered to predict the noiseless state $\mathbf{x}_0$. However, it's worth noting that different prediction objectives can be transformed into one another, we will delve into it in Section~\ref{sec:abalation}. Now, we consider the following alternative training loss weights:

\begin{itemize}
    \item Constant weighting. 
    $w_t = 1$. 
    Which treats different tasks as equally weighted and has been used in both discrete diffusion models~\cite{gu2022vqdiffusion, tang2022improved} and continuous diffusion models~\cite{cao2022survey}.
    \item SNR weighting. $w_t=\text{SNR}(t)$, where $\text{SNR}(t)=\alpha_t^2 / \sigma_t^2$. It's the most widely used weighting strategy~\cite{meng2022ondistillation,ho2022video,dhariwal2021adm,rombach2022ldm}. By combining with Equation~\ref{eq:diffusion}, we can find it's numerically equivalent to the constant weighting strategy when the predicting target is noise.
    \item Max-SNR-$\gamma$ weighting. $w_t=\max\{ \text{SNR}(t), \gamma \}$. This modification of SNR weighting is first proposed in ~\cite{salimans2022distillprogressive} to avoid a weight of zero with zero SNR steps. They set $\gamma=1$ as their default setting. However, the weights still concentrate on small noise levels.

    \item Min-SNR-$\gamma$ weighting. $w_t=\min\{ \text{SNR}(t), \gamma \}$. We propose this weighting strategy to avoid the model focusing too much on small noise levels.
    \item UGD optimization weighting. $w_t$ is optimized from Equation~\ref{eq:ugd-opt} in each timestep. Compared with the previous setting, this strategy changes during training.
\end{itemize}

\begin{figure}
    \centering
    \includegraphics[width=0.45\textwidth]{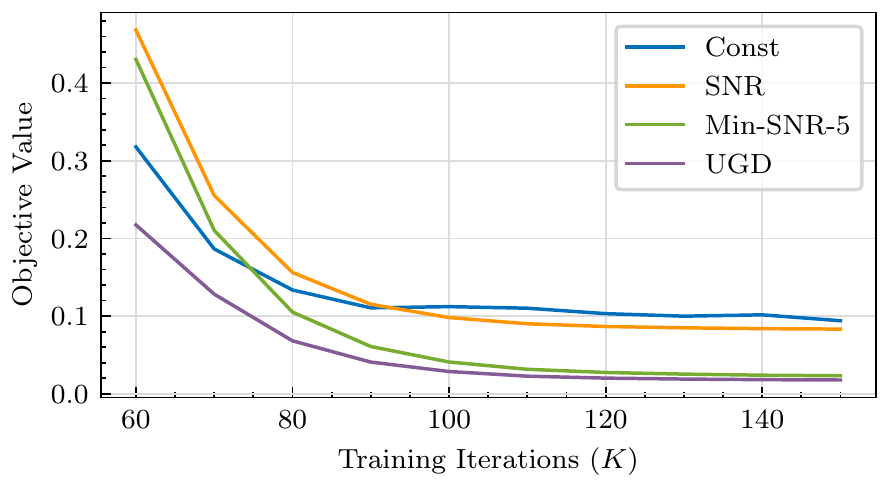}
    \vspace{-0.4cm}
    \caption{Comparison of the objective values in Equation~\ref{eq:reg-opt} on different weighting strategies.
    }
    \vspace{-0.3cm}
    \label{fig:object-value}
\end{figure}

First, we combine these weighting strategies into Equation~\ref{eq:reg-opt} to validate whether they are approach to the Pareto optimality state. As shown in Figure~\ref{fig:object-value}, the UGD optimization weighting strategy can achieve the lowest score on our optimization target. In addition, the Min-SNR-$\gamma$ weighting strategy is the closest to the optimum, demonstrating it has the property to optimize different timesteps simultaneously.

In the following section, we present experimental results to demonstrate the effectiveness of our Min-SNR-$\gamma$ weighting strategy in balancing diverse noise levels.
Our approach aims to achieve faster convergence and strong performance.

\section{Experiments}

In this section, we first provide an overview of the experimental setup. Subsequently, we conduct comprehensive ablation studies to show that our method is versatile and suitable for various prediction targets and network architectures. Finally, we compare our approach to the state-of-the-art methods across multiple image generation benchmarks, demonstrating not only its accelerated convergence but also its superior capability in generating high-quality images.

\subsection{Setup}\label{sec:implementation}

\noindent \textbf{Datasets.}
We perform experiments on both unconditional and conditional image generation using the CelebA dataset~\cite{liu2015CelebA} and the ImageNet dataset~\cite{deng2009imagenet}. 
The CelebA dataset, which comprises 162,770 human faces, is a widely-used resource for unconditional image generation studies. 
We follow ScoreSDE~\cite{2021Scoresde} for data pre-processing, which involves center cropping each image to a resolution of $140\times 140$ and then resizing it to $64\times 64$. 
For the class conditional image generation, we adopt the ImageNet dataset~\cite{deng2009imagenet} with a total of 1.3 million images from 1000 different classes. We test the performance on both $64\times 64$ and $256\times 256$ resolutions.

\noindent \textbf{Training Details.}\label{sec:training-detail}
For low resolution ($64\times 64$) image generation, we follow ADM~\cite{dhariwal2021adm} and directly train the diffusion model on the pixel-level. 
For high-resolution image generation, we utilize LDM~\cite{rombach2022ldm} approach by first compressing the images into latent space, then training a diffusion model to model the latent distributions. 
To obtain the latent for images, we employ VQ-VAE from Stable Diffusion\footnote{https://huggingface.co/stabilityai/sd-vae-ft-mse-original}, which encodes a high-resolution image ($256\times 256\times 3$) into $32\times 32 \times 4$ latent codes.

In our experiments, we employ both ViT and UNet as our diffusion model backbones. We adopt a vanilla ViT structure without any modifications~\cite{vit} as our default setting. we incorporate the timestep $t$ and class condition $\mathbf{c}$ as learnable input tokens to the model. 
Although further customization of the network structure may improve performance, our focus in this paper is to analyze the general properties of diffusion models. For the UNet structure, we follow ADM~\cite{dhariwal2021adm} and keep the FLOPs similar to the ViT-B model, which has $1.5\times$ parameters. Additional details can be found in the appendix.

For the diffusion settings, we use a cosine noise scheduler following the approach in~\cite{nichol2021iddpm,dhariwal2021adm}. The total number of timesteps is standardized to $T=1000$ across all datasets. We adopt AdamW~\cite{2014Adam,2018adamw} as our optimizer. 
For the CelebA dataset, we train our model for 500K iterations with a batch size of 128. During the first 5,000 iterations, we implement a linear warm-up and keep the learning rate at $1\times 10^{-4}$ for the remaining training. 
For the ImageNet dataset, the default learning rate is fixed at $1\times 10^{-4}$. 
The batch size is set to $1024$ for $64^2$ resolution and $256$ for $256^2$ resolution.

\noindent \textbf{Evaluation Settings.}\label{sec:evaluation}
To evaluate the performance of our models, we utilize an Exponential Moving Average (EMA) model with a rate of 0.9999. 
During the evaluation phase, we generate images with the Heun sampler from EDM~\cite{karras2022edm}. For conditional image generation, we also implement the classifier-free sampling strategy~\cite{ho2021classifierfree} to achieve better results. Finally, we measure the quality of the generated images using the FID score calculated on 50K images.

\subsection{Analysis of the Proposed Min-SNR-$\gamma$ }\label{sec:abalation}

\begin{figure*}[!h]
    \centering
    \includegraphics[width=0.33\textwidth]{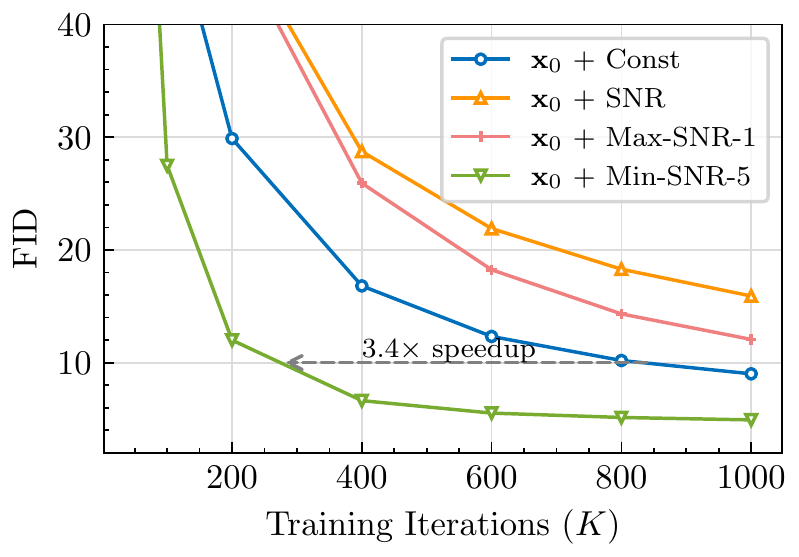}
    \includegraphics[width=0.33\textwidth]{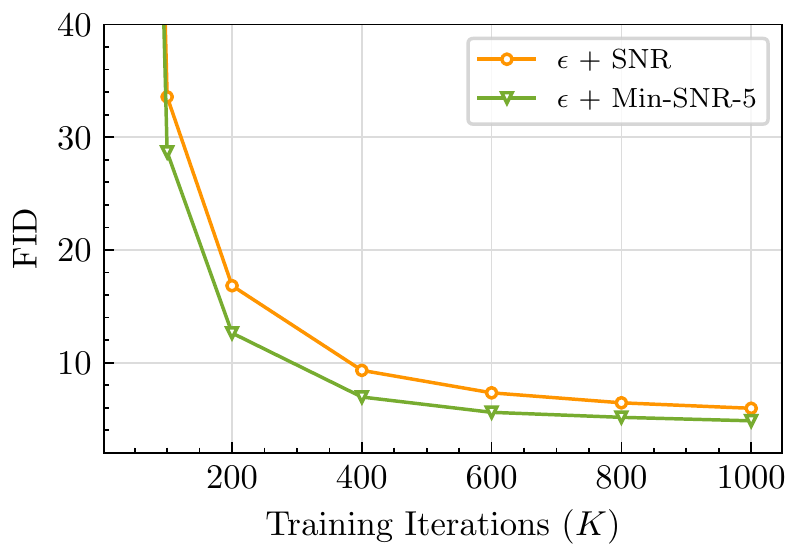}
    \includegraphics[width=0.33\textwidth]{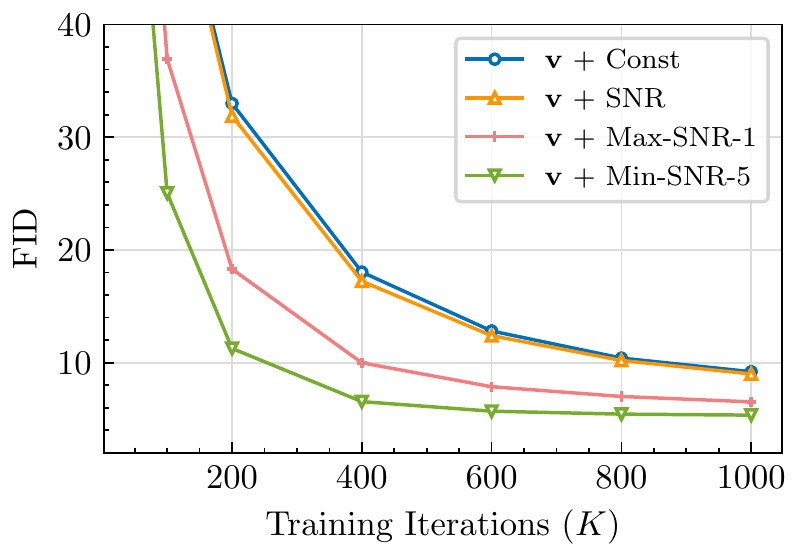}
    \vspace{-0.8cm}
    \caption{Comparing different loss weighting designs on predicting $\mathbf{x}_0$, $\mathbf{\epsilon}$, $\mathbf{v}$. Taking the neural network output as noise with const or Max-SNR-$\gamma$ strategy lead to divergence. Min-SNR-$\gamma$ strategy converges the fastest under all these settings.
    }
    \label{fig:abl-in256}
\end{figure*}

\begin{figure*}[!h]
    \centering
    \begin{subfigure}[b]{0.245\textwidth}
        \centering
        \includegraphics[width=\textwidth]{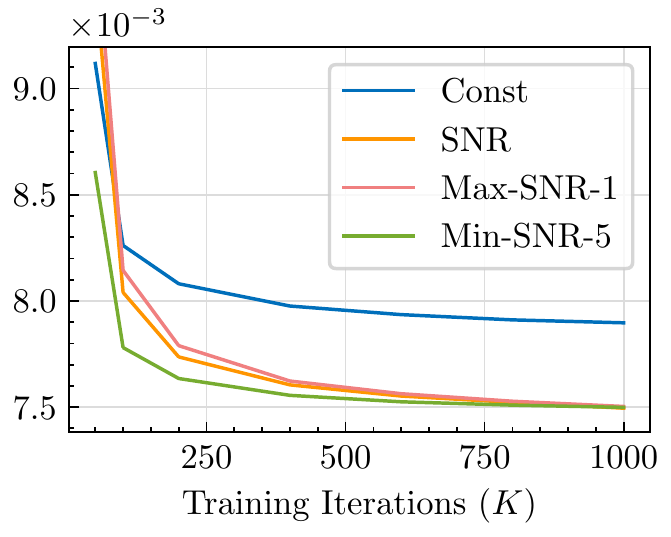}
    \end{subfigure}
    \hspace{-2mm}
    \begin{subfigure}[b]{0.245\textwidth}
        \centering
        \includegraphics[width=\textwidth]{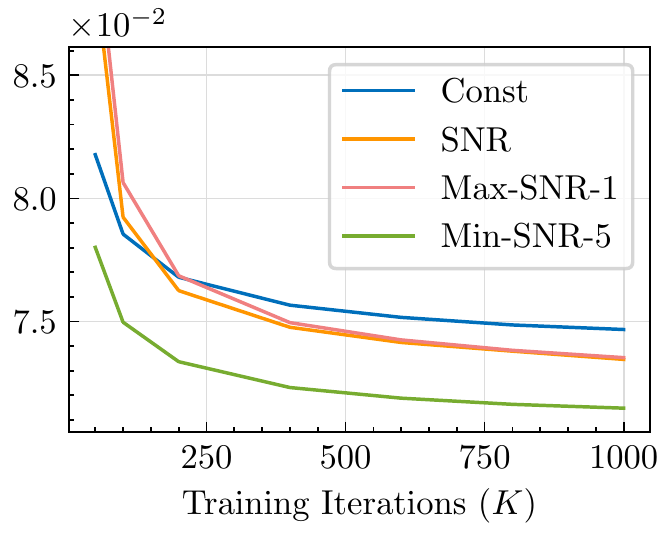}
    \end{subfigure}
    \hspace{-2mm}
    \begin{subfigure}[b]{0.245\textwidth}
        \centering
        \includegraphics[width=\textwidth]{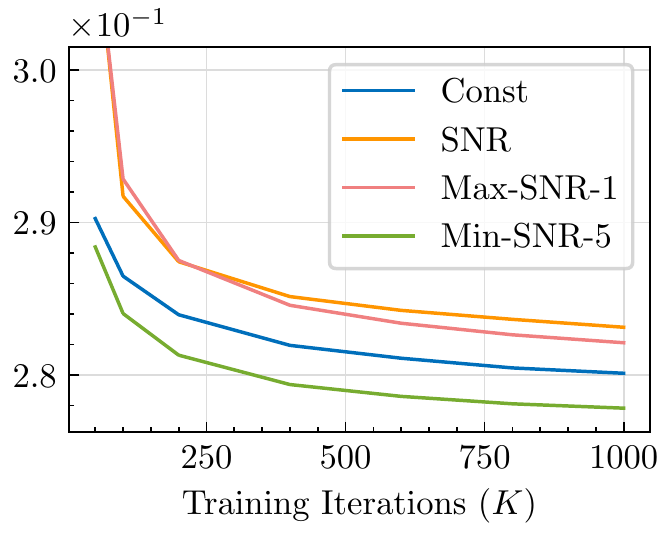}
    \end{subfigure}
    \hspace{-2mm}
    \begin{subfigure}[b]{0.25\textwidth}
        \centering
        \includegraphics[width=\textwidth]{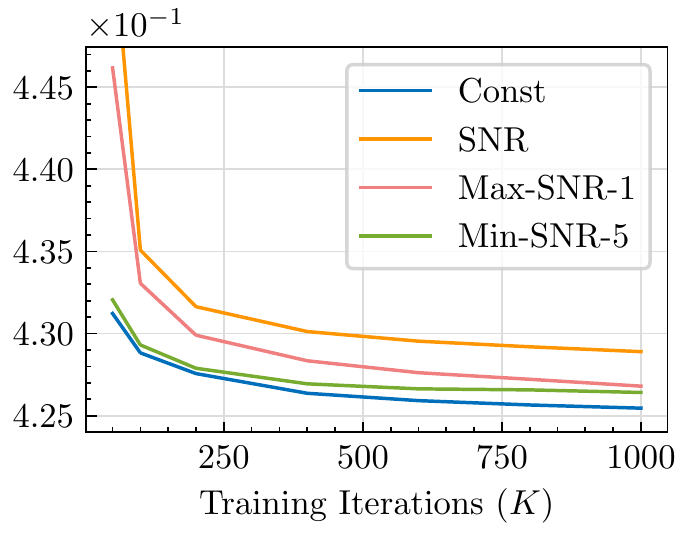}
    \end{subfigure}
    \vspace{-0.3cm}
    \caption{
    Unweighted loss in different ranges of timesteps. From left to right, each figure represents a specific range of timesteps: $[0, 100), [200, 300), [600, 700), [800, 900)$. The $y$-axis represents the Mean Squared Error (MSE), averaged over each range of timesteps.}
    \label{fig:viz-loss-diff-schedule}
\end{figure*}

\noindent \textbf{Comparison of Different Weighting Strategies.}
To demonstrate the significance of the loss weighting strategy, we conduct experiments with different loss weight settings for predicting $\mathbf{x}_0$. 
These settings include: 
1) constant weighting, where $w_t = 1$, 
2) SNR weighting, with $w_t = \text{SNR}(t)$, 
3) truncated SNR weighting, with $w_t = \max\{ \text{SNR}(t), \gamma \}$ (following~\cite{salimans2022distillprogressive} with a set value of $\gamma=1$), and 4) our proposed Min-SNR-$\gamma$ weighting strategy, with $w_t = \min\{ \text{SNR}(t), \gamma \}$, we set $\gamma=5$ as the default value.

The ViT-B serves as our default backbone and experiments are performed on ImageNet $256\times 256$. As illustrated in Figure~\ref{fig:abl-in256}, we observe that all results improve as the number of training iterations increases. However, our method demonstrates a significantly faster convergence compared to other methods. Specifically, it exhibits a $3.4\times$ speedup in reaching an FID score of $10$. It is worth mentioning that the SNR weighting strategy performed the worst, which could be due to its disproportionate focus on less noisy stages.

For a deeper understanding of the reasons behind the varying convergence rates, we analyzed their training loss at different noise levels. For a fair comparison, we exclude the loss weight term by only calculating $\lVert \mathbf{x}_0 - \mathbf{\hat{x}_{\theta}} \rVert_2^2$. 
Considering that the loss of different noise levels varies greatly, we calculate the loss in different bins and present the results in Figure~\ref{fig:viz-loss-diff-schedule}. 
The results show that while the constant weighting strategy is effective for high noise intensities, it performs poorly at low noise intensities. Conversely, the SNR weighting strategy exhibits the opposite behavior. 
In contrast, our proposed Min-SNR-$\gamma$ strategy achieves a lower training loss across all cases, and indicates quicker convergence through the FID metric.

\begin{figure*}[!h]
    \centering
    \includegraphics[width=\textwidth]{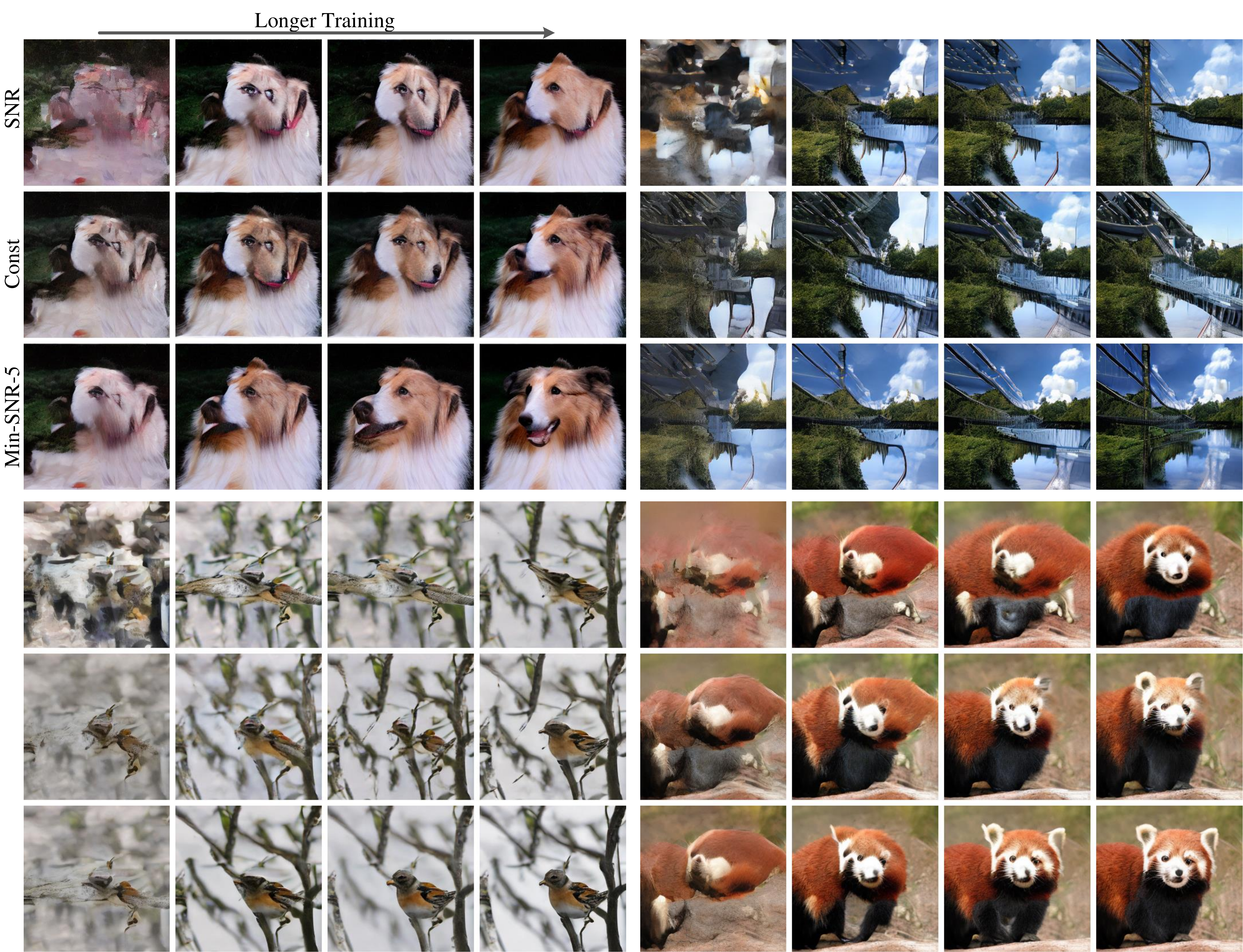}
    \caption{Qualitative comparison of the generation results from different weighting strategies on ImageNet-256 dataset. Images in each column are sampled from 50K, 200K, 400K, and 1M iterations. Our Min-SNR-5 strategy yields significant improvements in visual fidelity from the same iteration.}
    \label{fig:abl-loss-viz}
    \vspace{-5mm}
\end{figure*}

Furthermore, we present visual results in Figure~\ref{fig:abl-loss-viz} to demonstrate the fast convergence of the Min-SNR-$\gamma$ strategy. 
We apply the same random seed for noise to sample images from training iteration 50K, 200K, 400K, and 1M with different loss weight settings. Our results show that the Min-SNR-$\gamma$ strategy generates a clear object with only 200K iterations, which is significantly better in quality than the results obtained by other methods.

\noindent \textbf{Min-SNR-$\gamma$ for Different Prediction Targets.}
Instead of predicting the original signal $\mathbf{x}_0$ from the network, some recent works have employed alternative re-parameterizations, such as predicting noise $\epsilon$, or velocity $\mathbf{v}$~\cite{salimans2022distillprogressive}. To verify the applicability of our weighting strategy to these prediction targets, we conduct experiments comparing the four aforementioned weighting strategies across these different re-parameterizations. 

As we discussed in Section~\ref{sec:different-loss-strategy}, predicting noise $\epsilon$ is mathematically equivalent to predicting $\mathbf{x}_0$ by intrinsically involving Signal-to-Noise Ratio as a weight factor, thus we divide the SNR term in practice. For example, the Min-SNR-$\gamma$ strategy in predicting noise can be expressed as $w_t = \frac{\min\{ \text{SNR}(t), \gamma \}}{\text{SNR}(t)} = \min\{ \frac{\gamma}{\text{SNR}(t)}, 1 \}$. And the SNR strategy in predicting noise is equivalent to a ``constant strategy''. For simplicity and consistency, we still refer to them as Min-SNR-$\gamma$ and SNR strategies. Similarly, we can derive that when predicting velocity $\mathbf{v}$, the loss weight factor must be divided by $(\text{SNR}+1)$. These strategies are still referred to by their original names for ease of reference.

We conduct experiments on these two variants and present the results in Figure~\ref{fig:abl-in256}. Taking the neural network output as noise with const or Max-SNR-$\gamma$ setting leads to divergence. Meanwhile, our proposed Min-SNR-$\gamma$ strategy converges faster than other loss weighting strategies for both prediction noise and predicting velocity. These demonstrate that balancing the loss weights for different timesteps is intrinsic, independent of any re-parameterization.

\noindent \textbf{Min-SNR-$\gamma$ on Different Network Architectures.}
The Min-SNR-$\gamma$ strategy is versatile and robust for different prediction targets and network structures. We conduct experiments on the widely used UNet and keep the number of parameters close to the ViT-B model. For each experiment, models were trained for 1 million iterations and their FID scores were calculated at multiple intervals. The results in Table~\ref{tab:abl-unet} indicate that the Min-SNR-$\gamma$ strategy converges significantly faster than the baseline and provides better performance for both predicting $\mathbf{x}_0$ and predicting noise.

\begin{table}[t]
\begin{center}
\setlength{\tabcolsep}{1.5mm}
\begin{tabular}{l c c c c c}
\toprule
Training Iterations & 200K & 400K & 600K & 800K & 1M \\
\midrule
Baseline ($\mathbf{x}_0$) & 25.93 & 15.41 & 11.54 & 9.52 & 8.33 \\
+ Min-SNR-$5$ & \textbf{7.99} & \textbf{5.34} & \textbf{4.69} & \textbf{4.41} & \textbf{4.28} \\
\midrule
Baseline ($\epsilon$) & 8.55 & 5.43 & 4.64 & 4.35 & 4.21 \\
+ Min-SNR-$5$ & \textbf{7.32} & \textbf{4.98} & \textbf{4.48} & \textbf{4.24} & \textbf{4.14} \\
\bottomrule
\end{tabular}
\end{center}
\vspace{-5mm}
\caption{
Ablation studies on the UNet backbone. Whether the network predicts $\mathbf{x}_0$ or $\epsilon$, the Min-SNR-5 weighting design converges faster and achieves better FID score.
}
\label{tab:abl-unet}
\vspace{-5mm}
\end{table}

\noindent \textbf{Robustness Analysis.}
Our approach utilizes a single hyperparameter, $\gamma$, as the truncate value. To assess its robustness, we conducted a thorough robustness analysis in various settings. Our experiments were performed on the ImageNet-256 dataset using the ViT-B model and the prediction target of the network is $\mathbf{x}_0$. We varied the truncate value $\gamma$ by setting it to 1, 5, 10, and 20 and evaluated their performance. The results are shown in Table~\ref{tab:abl-diff-k}. We find there are only minor variations in the FID score when $\gamma$ is smaller than 20. Additionally, we conducted more experiments by modifying the predicting target to the noise $\epsilon$, and modifying the network structure to UNet. We find that the results were also consistently stable. Our results indicate that good performance can usually be achieved when $\gamma$ is set to 5, making it the established default setting.

\begin{table}[t]
\begin{center}
\setlength{\tabcolsep}{2mm}
\begin{tabular}{l c c c c }
\toprule
$\gamma$ & 1 & 5 & 10 & 20  \\
\midrule
ViT ($\mathbf{x}_0$) & 4.98 & \textbf{4.92} & 5.34 & 5.45  \\
ViT ($\epsilon$)  & 4.89 & \textbf{4.84} & 4.94 & 5.41  \\
UNet ($\mathbf{x}_0$) & 4.49 & \textbf{4.28} & 4.32 & 4.37  \\
UNet ($\epsilon$) & 4.30 & 4.14 & 4.14 & \textbf{4.12}  \\
\bottomrule
\end{tabular}
\end{center}
\vspace{-5mm}
\caption{ 
Ablation study on $\gamma$.
The results are robust to the hyper-parameter $\gamma$ in different settings.}
\label{tab:abl-diff-k}
\vspace{-3mm}
\end{table}

\begin{table}[t]
\begin{center}
\begin{tabular}{l c c}
\toprule
Method & \#Params & FID \\
\midrule
DDIM~\cite{ddim} & 79M & 3.26 \\
Soft Truncation~\cite{ddim} & 62M & 1.90 \\
\textbf{Our UNet} & 59M & \textbf{1.60} \\
\midrule
U-ViT-Small~\cite{bao2022uvit} & 44M &   2.87 \\
\textbf{ViT-Small (ours)} & 43M &   \textbf{2.14} \\
\bottomrule
\end{tabular}
\end{center}
\vspace{-5mm}
\caption{FID results of unconditional image generation on CelebA $64\times 64$~\cite{liu2015CelebA}. 
We conduct experiments with both UNet and ViT backbone.
}
\label{tab:CelebA64}
\vspace{-3mm}
\end{table}

\subsection{Comparison with state-of-the-art Methods}\label{sec:compare-with-sota}
\noindent\textbf{CelebA-64.}
We conduct experiments on the CelebA $64\times 64$ dataset for unconditional image generation. Both UNet and ViT are used as our backbones and are trained for 500K iterations. During the evaluation, we use the EDM sampler~\cite{karras2022edm} to generate 50K samples and calculate the FID score. The results are summarized in Table~\ref{tab:CelebA64}. Our ViT-Small~\cite{vit} model outperforms previous ViT-based models with an FID score of 2.14. It is worth mentioning that no modifications are made to the naive network structure, demonstrating that the results could still be improved further. Meanwhile, our method using the UNet~\cite{dhariwal2021adm} structure achieves an even better FID score of 1.60, outperforming previous UNet methods.

\noindent\textbf{ImageNet-64.}
We also validate our method on class-conditional image generation on the ImageNet $64\times 64$ dataset. 
During training, the class label is dropped with the probability $0.15$ for classifier-free inference~\cite{ho2021classifierfree}. 
The model is trained for 800K iterations and images are synthesized using classifier-free guidance with a scale of $\text{cfg}=1.5$ and the EDM sampler for image generation. 
For a fair comparison, we adopt a 21-layer ViT-Large model without additional architecture designs, which has a similar number of parameters to U-ViT-Large~\cite{bao2022uvit}. 
The results presented in Table~\ref{tab:in64} show that our method achieves an FID score of 2.28, significantly improving upon the U-ViT-Large model.

\begin{table}[t]
\begin{center}
\begin{tabular}{l c c }
\toprule
Method & \#Params & FID \\
\midrule
BigGAN-deep~\cite{brock2018biggan} & {} & 4.06 \\
StyleGAN-XL~\cite{sauer2022styleganxl} & {} & 1.51 \\
\midrule
IDDPM (small)~\cite{nichol2021iddpm} & 100M & 6.92 \\
IDDPM (large)~\cite{nichol2021iddpm} & 270M &  2.92 \\
CDM~\cite{ho2022cdm} & {} &  1.48 \\
ADM~\cite{dhariwal2021adm} & 296M &  2.61 \\
\midrule
U-ViT-Mid~\cite{bao2022uvit} & 131M &   5.85 \\
U-ViT-Large~\cite{bao2022uvit} & 287M &   4.26 \\
\textbf{ViT-L (ours)} & 269M &   \textbf{2.28} \\
\bottomrule
\end{tabular}
\end{center}
\vspace{-5mm}
\caption{
FID results on ImageNet $64\times 64$. We conduct experiments using the ViT-L backbone which significantly improves upon previous methods.}
\label{tab:in64}
\end{table}

\begin{table}[t]
\begin{center}
\begin{tabular}{ l c c }
\toprule
Method & \#Params & FID \\
\midrule
BigGAN-deep~\cite{brock2018biggan} & 340M & 6.95 \\
StyleGAN-XL~\cite{sauer2022styleganxl} & {} & 2.30 \\
\midrule
Improved VQ-Diffusion~\cite{gu2022vqdiffusion} & 460M & 4.83 \\
\midrule
IDDPM~\cite{nichol2021iddpm} & 270M &  12.26 \\
CDM~\cite{ho2022cdm} & {} &  4.88 \\
ADM~\cite{dhariwal2021adm} & 554M &  10.94 \\
ADM-U~\cite{dhariwal2021adm} & 608M &  7.49 \\
ADM-G~\cite{dhariwal2021adm} & 554M &  4.59 \\
ADM-U, ADM-G~\cite{dhariwal2021adm} & 608M &  3.94 \\
LDM~\cite{rombach2022ldm} & 400M & 3.60 \\
\textbf{UNet (ours)} & 395M & \textbf{2.81}$^\dagger$ \\
\midrule
U-ViT-L~\cite{bao2022uvit} & 287M &   3.40 \\
DiT-XL-2~\cite{peebles2022dit} & 675M &   9.62 \\
DiT-XL-2 (cfg=1.50)~\cite{peebles2022dit}  & 675M &   2.27 \\
{ViT-XL (ours)} & 451M &   8.10 \\
\textbf{ViT-XL (ours, cfg=1.50)} & 451M &   \textbf{2.06} \\
\bottomrule
\end{tabular}
\end{center}
\vspace{-5mm}
\caption{
FID results on ImageNet $256\times 256$. $^\dagger$ denotes only train 1.4M iterations. Our model with a ViT-XL backbone achieves a new record FID score of 2.06. 
}
\label{tab:in256}
\vspace{-3mm}
\end{table}

\noindent\textbf{ImageNet-256.}
We also apply diffusion models for higher-resolution image generation on the ImageNet $256\times 256$ benchmark. To enhance training efficiency, we first compress $256\times 256 \times 3$ images into $32\times 32 \times 4$ latent codes using the encoder from LDM~\cite{rombach2022ldm}. During the sampling process, we employ the EDM sampler and the classifier-free guidance to generate images. 
The FID comparison is presented in Table~\ref{tab:in256}. 
Under the setting of predicting $\epsilon$ with Min-SNR-5, our ViT-XL model achieves the FID of $2.08$ for only 2.1M iterations, which is $3.3\times$ faster than DiT and outperforms the previous state-of-the-art FID record of $2.27$. 
Moreover, with longer training (about 7M iterations as in~\cite{peebles2022dit}), we are able to achieve the FID score of 2.06 by predicting $\mathbf{x}_0$ with Min-SNR-5.
Our UNet-based model with 395M parameters is trained for about 1.4M iterations and achieves FID score of 2.81.

\section{Conclusion}
In this paper, we point out that the conflicting optimization directions between different timesteps may cause slow convergence in diffusion training. To address it, we regard the diffusion training process as a multi-task learning problem and introduce a novel weighting strategy, named Min-SNR-$\gamma$, to effectively balance different timesteps. Experiments demonstrate our method can boost diffusion training several times faster, and achieves the state-of-the-art FID score on ImageNet-256 dataset.

\section*{Acknowledgments}
We sincerely thank Yixuan Wei, Zheng Zhang, and Stephen Lin for helpful discussion.
This research was partly supported by the National Key Research \& Development Plan of China (No. 2018AAA0100104), the National Science Foundation of China (62125602, 62076063).

{\small
\bibliographystyle{ieee_fullname}
\bibliography{egbib}
}

\appendix

In the appendix, we first provide the proof of Theorem 1 in Section~\ref{sec:ugd}.
Then we derive the relationship between loss weights of different predicting targets in Section~\ref{sec:transformation}. 
In Section~\ref{sec:hyper-parameter}, we provide more details on the network architecture, training and sampling settings. Finally, we present more visual results in Section~\ref{sec:addtional-results}.

\section{Proof for Theorem 1}~\label{sec:ugd}
First, we introduce the Pareto Optimality mentioned in the paper.
Assume the loss for each task is $\mathcal{L}^t(\theta), t \in \{ 1, 2,\ldots, T\}$ and the respective gradient to $\theta$ is $\nabla_\theta \mathcal{L}^t(\theta)$.
For simplicity, we denote $\mathcal{L}^t(\theta)$ as $\mathcal{L}^t$.
If we treat each task with equal importance, we assume each loss item $\mathcal{L}^1, \mathcal{L}^2, \ldots, \mathcal{L}^T$ is decreasing or kept the same.
There exists one point $\theta^*$ where any change of the point will leads to the increase of one loss item.
We call the point $\theta^*$ ``Pareto Optimality''.
In other words, we cannot sacrifice one task for another task's improvement. 
To reach Pareto Optimality, we need to find an update direction $\delta$ which meet:
\begin{align}
    \left\{ 
    \begin{array}{cc}
         \left\langle \nabla_\theta \mathcal{L}_{\theta}^1, \delta \right\rangle & \leq 0  \\
         \left\langle \nabla_\theta \mathcal{L}_{\theta}^2, \delta \right\rangle & \leq 0  \\
         \vdots & {}  \\
         \left\langle \nabla_\theta \mathcal{L}_{\theta}^T, \delta \right\rangle & \leq 0  \\
    \end{array}
    \right.
\end{align}
$\left\langle \cdot,\cdot \right\rangle$ denotes the inner product of two vectors.
It is worth noting that $\delta=0$ satisfies all the above inequalities.
We care more about the non-zero solution and adopt it for updating the network parameter $\theta$. 
If the non-zero point does not exist, it may already achieve the ``Pareto Optimality'', which is referred as ``Pareto Stationary''.

For simplicity, we denote the gradient for each loss item $\nabla_\theta\mathcal{L}^t$ as $\mathbf{g}_t$. Suppose we have a gradient vector $\mathbf{u}$ to satisfy that all $\left\langle \mathbf{g}_t, \mathbf{u} \right\rangle \geq 0, t\in \{1,2,\ldots,T \}$. 
Then $-\mathbf{u}$ is the updating direction ensuring a lower loss for each task.

As proposed in~\cite{kexuefm8896}, $\left\langle \mathbf{g}_t, \mathbf{u} \right\rangle \geq 0, \forall t \in \{ 1, 2, \ldots, T \}$ is equivalent to $\min_t \left\langle \mathbf{g}_t, \mathbf{u} \right\rangle \geq 0$.
And it could be achieved when the minimal value of $\left\langle \mathbf{g}_t, \mathbf{u} \right\rangle$ is maximized.
Thus the problem is further converted to:
\begin{align*}
    \max_{\mathbf{u}} \min_t \left\langle \mathbf{g}_t, \mathbf{u} \right\rangle
\end{align*}
There is no constraint for the vector $\mathbf{u}$, so it may become infinity and make the updating unstable.
To avoid it, we add a regularization term to it
\begin{align}\label{eq:objective-function}
    \max_{\mathbf{u}} \min_t \left\langle \mathbf{g}_t, \mathbf{u} \right\rangle - \frac{1}{2}\lVert \mathbf{u}  \rVert _2^2.
\end{align}
And notice that the $\max$ function ensures the value is always greater than or equal to a specific value $\mathbf{u}=0$. 
\begin{align*}
 &\max_{\mathbf{u}} \min_t \left\langle \mathbf{g}_t, \mathbf{u} \right\rangle - \frac{1}{2}\lVert \mathbf{u}  \rVert_2^2 \\
 \geq   & \left.  \min_t \left\langle \mathbf{g}_t, \mathbf{u}
 \right\rangle - \frac{1}{2}\lVert \mathbf{u}  \rVert _2^2 \right| _{\mathbf{u}=0} \\
 =&~ 0,
 \end{align*}
 which also means $\max_{\mathbf{u}} \min_t \left\langle \mathbf{g}_t, \mathbf{u} \right\rangle \geq \frac{1}{2}\lVert \mathbf{u}  \rVert_2^2 \geq 0 $. Therefore, the solution of Equation~\ref{eq:objective-function} satisfies our optimization goal of $\left\langle \mathbf{g}_t, \mathbf{u} \right\rangle \geq 0, \forall t \in \{ 1, 2, \ldots, T \}$.

We define $\mathcal{C}^T$ as a set of $n$-dimensional variables
\begin{align}
    \mathcal{C}^T = \left\{ (w_1, w_2, \ldots, w_T)  | w_1, w_2, \ldots, w_T \geq 0, \sum_{t=1}^T w_t = 1 \right\},
\end{align}
It is easy to verify that 
\begin{align}
    \min_t \left\langle \mathbf{g}_t,  \mathbf{u}\right\rangle =  \min_{w \in \mathcal{C}^T} \left\langle \sum_t w_t  \mathbf{g}_t, \mathbf{u} \right\rangle.
\end{align}

We can also verify the above function is concave with respect to $\mathbf{u}$ and $\alpha$. According to Von Neumann's Minmax theorem~\cite{von1947theory}, the objective with regularization in Equation~\ref{eq:objective-function} is equivalent to 
\begin{align}
    & \max_{\mathbf{u}} \min_{w \in \mathcal{C}^T} \left\{\left\langle \sum_t w_t  \mathbf{g}_t, \mathbf{u} \right\rangle - \frac{1}{2}\lVert \mathbf{u}  \rVert_2^2\right\} \\
    = & \min_{w \in \mathcal{C}^T} \max_{\mathbf{u}} \left\{ \left\langle \sum_t w_t  \mathbf{g}_t, \mathbf{u} \right\rangle - \frac{1}{2}\lVert \mathbf{u}  \rVert_2^2 \right\} \\
    = & \left. \min_{w \in \mathcal{C}^T} \left\{  \left\langle \sum_t w_t  \mathbf{g}_t, \mathbf{u} \right\rangle - \frac{1}{2}\lVert \mathbf{u}  \rVert_2^2 \right\} \right|_{\mathbf{u}=\frac{1}{2} \sum_t w_t  \mathbf{g}_t} \\
    = & \min_{w \in \mathcal{C}^T} \frac{1}{2} \left\lVert \sum_t w_t  \mathbf{g}_t \right\rVert_2^2.
\end{align}
Finally, we achieved Theorem 1 in the main paper.

\section{Relationship between Different Targets}\label{sec:transformation}
The most common predicting target is in $\epsilon$-space.
Loss for prediction in $\mathbf{x}_0$-space and $\epsilon$-space can be transformed by the SNR loss weight.
\begin{align*}
    \mathcal{L}_\theta &= \left\lVert 
\epsilon - \hat{\epsilon}_\theta (\mathbf{x}_t)  \right\rVert_2^2 \\
&= \left\lVert \frac{1}{\sigma_t} (\mathbf{x}_t - \alpha_t \mathbf{x}_0 ) - \frac{1}{\sigma_t} (\mathbf{x}_t - \alpha_t \hat{\mathbf{x}}_\theta (\mathbf{x}_t)) \right\rVert_2^2 \\
&= \frac{\alpha_t^2}{\sigma_t^2} \left\lVert 
\mathbf{x}_0 - \hat{\mathbf{x}}_\theta (\mathbf{x}_t))  \right\rVert_2^2 \\
&= \text{SNR}(t) \left\lVert 
\mathbf{x}_0 - \hat{\mathbf{x}}_\theta (\mathbf{x}_t))  \right\rVert_2^2,
\end{align*}
where $\hat{\epsilon}_\theta$ is the network to predict the noise and $ \hat{\mathbf{x}}_\theta$ is to predict the clean data.

Prediction target $\mathbf{v} = \alpha_t \epsilon - \sigma_t \mathbf{x}_0$ is proposed in~\cite{salimans2022distillprogressive}, we can derive the related loss
\begin{align*}
    \mathcal{L}_\theta &= \left\lVert \mathbf{v}_t - \mathbf{v}_\theta (\mathbf{x}_t) \right\rVert_2^2 \\
    &= \left\lVert\left( \alpha_t \epsilon - \sigma_t \mathbf{x}_0\right) - \left( \alpha_t \hat{\epsilon}_\theta (\mathbf{x}_t) - \sigma_t \hat{\mathbf{x}}_\theta(\mathbf{x}_t)\right) \right\rVert_2^2 \\
    &= \left\lVert\alpha_t\left(  \epsilon - \hat{\epsilon}_\theta (\mathbf{x}_t) \right)  - \sigma_t \left(   \mathbf{x}_0 -  \hat{\mathbf{x}}_\theta(\mathbf{x}_t)\right) \right\rVert_2^2 \\
    &= \left\lVert \alpha_t \frac{\alpha_t}{\sigma_t}\left( \hat{\mathbf{x}}_\theta(\mathbf{x}_t)   - \mathbf{x}_0 \right)- \sigma_t \left(   \mathbf{x}_0 -  \hat{\mathbf{x}}_\theta(\mathbf{x}_t)\right) \right\rVert_2^2 \\
    &= \left\lVert \frac{\alpha_t^2 + \sigma_t^2}{\sigma_t} \left(   \mathbf{x}_0 -  \hat{\mathbf{x}}_\theta(\mathbf{x}_t)\right) \right\rVert_2^2 \\
    &= \frac{1}{\sigma_t^2} \left\lVert  \left(   \mathbf{x}_0 -  \hat{\mathbf{x}}_\theta(\mathbf{x}_t)\right) \right\rVert_2^2 \\
    &= \frac{\alpha_t^2 + \sigma_t^2}{\sigma_t^2} \left\lVert  \left(   \mathbf{x}_0 -  \hat{\mathbf{x}}_\theta(\mathbf{x}_t)\right) \right\rVert_2^2 \\
    &= (\text{SNR}(t)+1) \left\lVert  \left(   \mathbf{x}_0 -  \hat{\mathbf{x}}_\theta(\mathbf{x}_t)\right) \right\rVert_2^2 \\
\end{align*}

\section{Hyper-parameter}\label{sec:hyper-parameter}
Here we list more details about the architecture, training and evaluation setting.

\subsection{Architecture Settings}
The ViT setting adopted in the paper are as follows,
\begin{table}[!h]
\begin{center}
\setlength{\tabcolsep}{1.5mm}
\begin{tabular}{l c c c c}
\toprule
Model & Layers & Hidden Size &  Heads & Params \\
\midrule
ViT-Small & 13 & 512 & 6 & 43M \\
ViT-Base & 12 & 768 & 12 & 88M \\
ViT-Large   & 21 & 1024 & 16 & 269M \\
ViT-XL   & 28 & 1152 & 16 & 451M \\
\bottomrule
\end{tabular}
\end{center}
\caption{
Configurations of our used ViTs.
}
\label{tab:config-vit}
\end{table}

We use ViT-Small for face generation on CelebA $64\times 64$. Besides, we adopt ViT-Base as the default backbone for the ablation study.
To make relative fair comparison with U-ViT, we use a 21-layer ViT-Large for ImageNet $64\times 64$ benchmark.
To compare with former state-of-the-art method DiT~\cite{peebles2022dit} on ImageNet $256\times 256$, we adopt the similar setting ViT-XL with the same depth, hidden size, and patch size.

In the paper, we also evaluate our method's robustness to model architectures using the UNet backbone.
For ablation study, we adjust the setting based on ADM~\cite{dhariwal2021adm} to make the parameters and FLOPs close to ViT-B. The setting is 
\begin{itemize}
    \item Base channels: 192
    \item Channel multipliers: 1, 2, 2, 2
    \item Residual blocks per resolution: 3
    \item Attention resolutions: 8, 16
    \item Attention heads: 4
\end{itemize}

We also conduct experiments with the same architecture (296M) in ADM~\cite{dhariwal2021adm} on ImageNet $64\times 64$.
After 900K training iterations with batch size 1024, it could achieve an FID score of 2.11.

For high resolution generation on ImageNet $256\times 256$.
We use the 395M setting from LDM~\cite{rombach2022ldm}, which operates on the $32\times 32 \times 4$ latent space.

\subsection{Training Settings}
The training iterations and learning rate have been reported in the paper.
We use AdamW~\cite{2018adamw,2014Adam} as our default optimizer.
$(\beta_1, \beta_2)$ is set to $(0.9,0.999)$ for UNet backbone.
Following~\cite{bao2022uvit}, we set $(\beta_1, \beta_2)$ to $(0.99,0.99)$ for ViT backbone.

\subsection{Sampling Settings}
If not otherwise specified, we only use EDM's~\cite{karras2022edm} Heun sampler.
We only adjust the sampling steps for better results.
For ablation study with ViT-B and UNet, we set the number of steps to 30.
For ImageNet $64 \times 64$ in Table~4, the number of steps is set to 20.
For ImageNet $256\times 256$ in Table~5, the number of sampling steps is set to 50.

\section{Additional Results}\label{sec:addtional-results}

\subsection{Ablation Study on Pixel Space}
In the paper, most of the ablation study is conducted on ImageNet $256\times 256$'s latent space.
Here, we present the results on ImageNet $64\times 64$ pixel space.
We adopt a ViT-B model as our backbone and train the diffusion model for 800K iterations with batch size 512. 
Our predicting targets are $\mathbf{x}_0$ and $\epsilon$ and they are equipped with our proposed simple Min-SNR-$\gamma$ loss weight ($\gamma=5$).
We adopt the pre-trained noisy classifier at $64\times 64$ from ADM~\cite{dhariwal2021adm} as conditional guidance.
We can see that the loss weighting strategy contributes to the faster convergence for both $\mathbf{x}_0$ and $\epsilon$.

\begin{figure}[!h]
    \centering
    \includegraphics[width=0.42\textwidth]{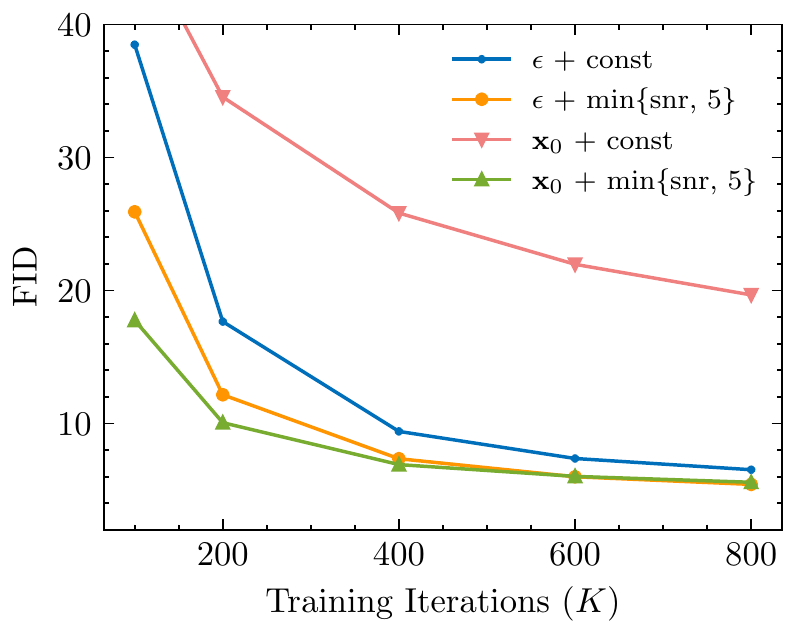}
    \caption{Ablate loss weight design in pixel space (ImageNet $64\times 64$). We adopt DPM Solver~\cite{Lu2022DPMSolverAF} to sample $50k$ images to calculate the FID score with classifier guidance.}
    \label{fig:abl-in64}
\end{figure}

\subsubsection{Min-SNR-$\gamma$ on EDM}
\begin{figure}[!h]
    \centering
    \includegraphics{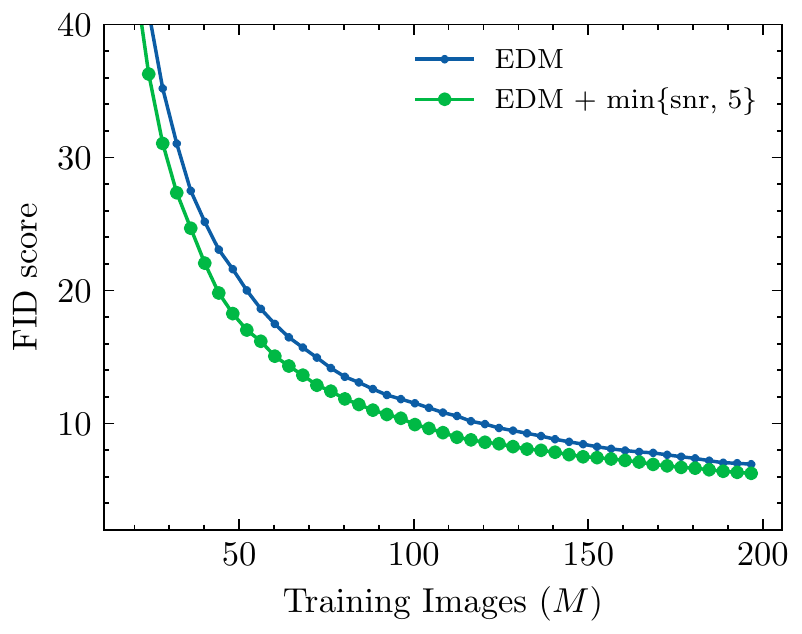}
    \caption{Effect of Min-SNR-$\gamma$ on EDM~\cite{karras2022edm}.}
    \label{fig:edm-min-snr}
\end{figure}

We also apply our Min-SNR-$\gamma$ weighting strategy on the SoTA ``denoiser'' framework EDM.
We find that our strategy can also help converge faster in such framework in Figure~\ref{fig:edm-min-snr}. The specific implementation is to multiply $\frac{\min \{ \text{SNR}, 5 \}}{\text{SNR}}$ in \texttt{EDMLoss} from official code\footnote{https://github.com/NVlabs/edm.git}.
We keep the same setting as official ImageNet-64 training setting, including batch size and optimizer. Due to the limit of compute budget, we did not train the model as long as that in EDM~\cite{karras2022edm} (about 2k epochs on ImageNet). 
We use $2^{nd}$ Heun approach with 18 steps (NFE=35).
The curve in Figure~\ref{fig:edm-min-snr} reflects the FID's changing with training images.

\subsection{Visual Results on Different Datasets}
\begin{figure*}
    \centering
    \includegraphics[width=0.95\textwidth]{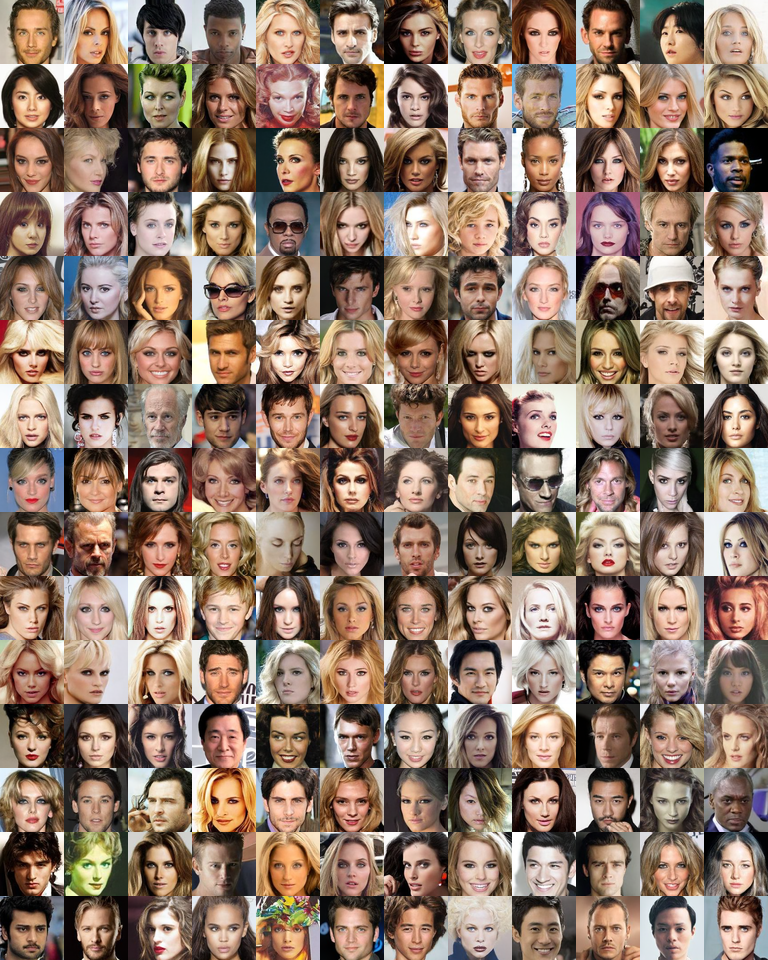}
    \caption{Additional generated samples on CelebA $64\times 64$. The samples are from UNet backbone with 1.60 FID.}
    \label{fig:CelebA64}
\end{figure*}

\begin{figure*}
    \centering
    \includegraphics[width=0.95\textwidth]{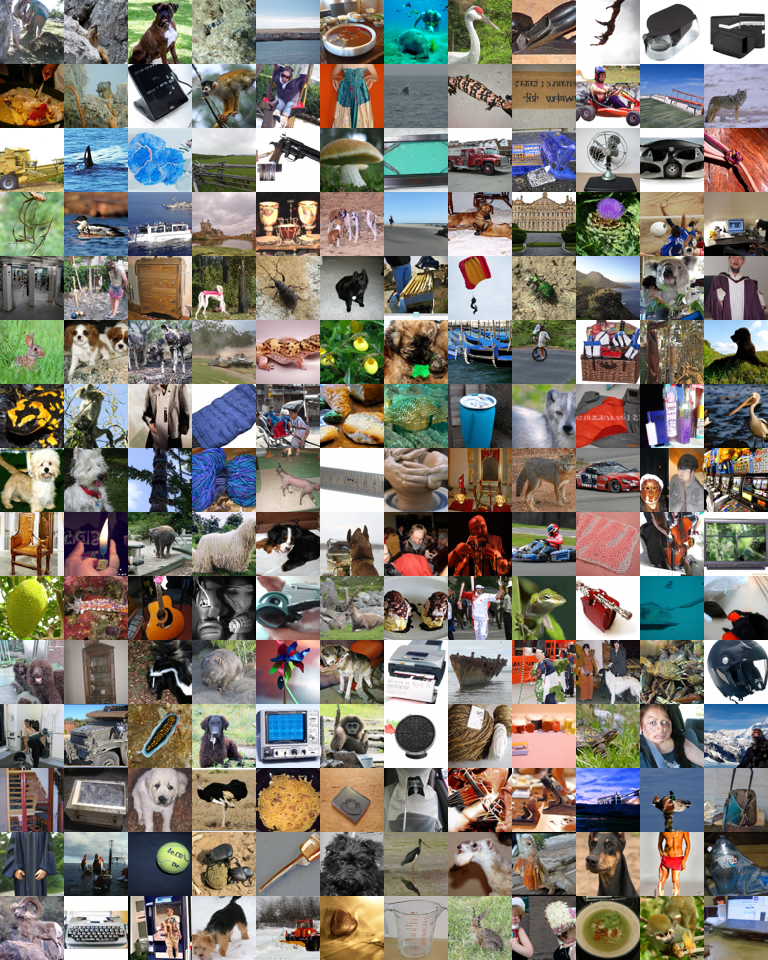}
    \caption{Additional generated samples on ImageNet $64\times 64$. The samples are from ViT backbone with 2.28 FID.}
    \label{fig:vit_tn64}
\end{figure*}

\begin{figure*}
    \centering
    \includegraphics[width=0.95\textwidth]{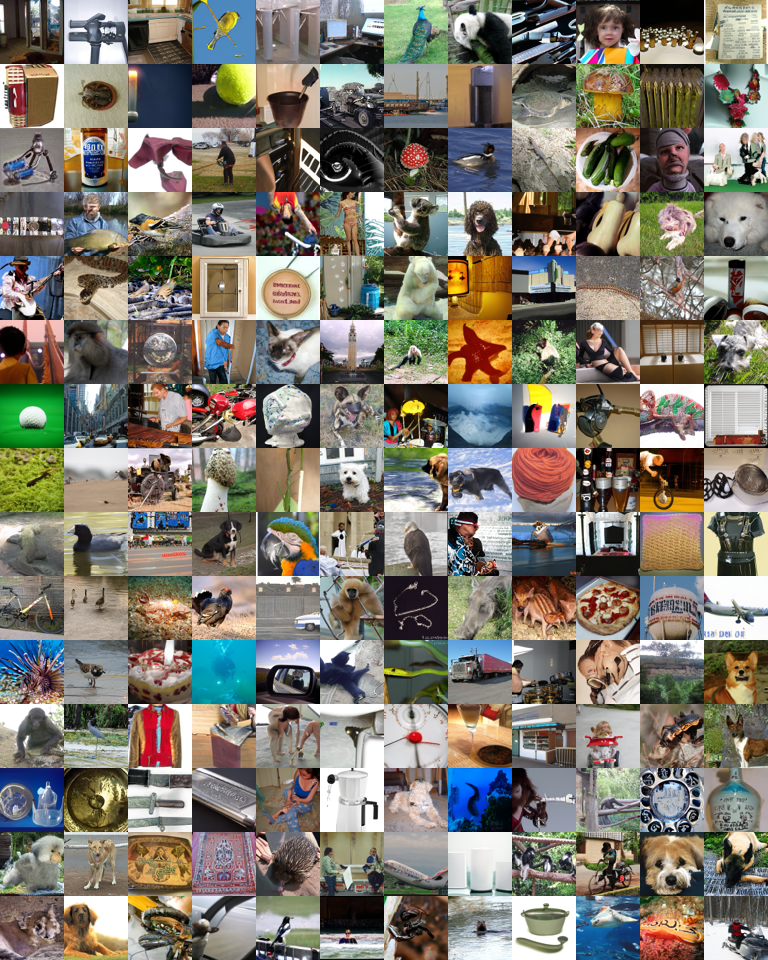}
    \caption{Additional generated samples on ImageNet $64\times 64$. The samples are from UNet backbone with 2.14 FID.}
    \label{fig:unet_in64}
\end{figure*}

\begin{figure*}
    \centering
    \includegraphics[width=0.95\textwidth]{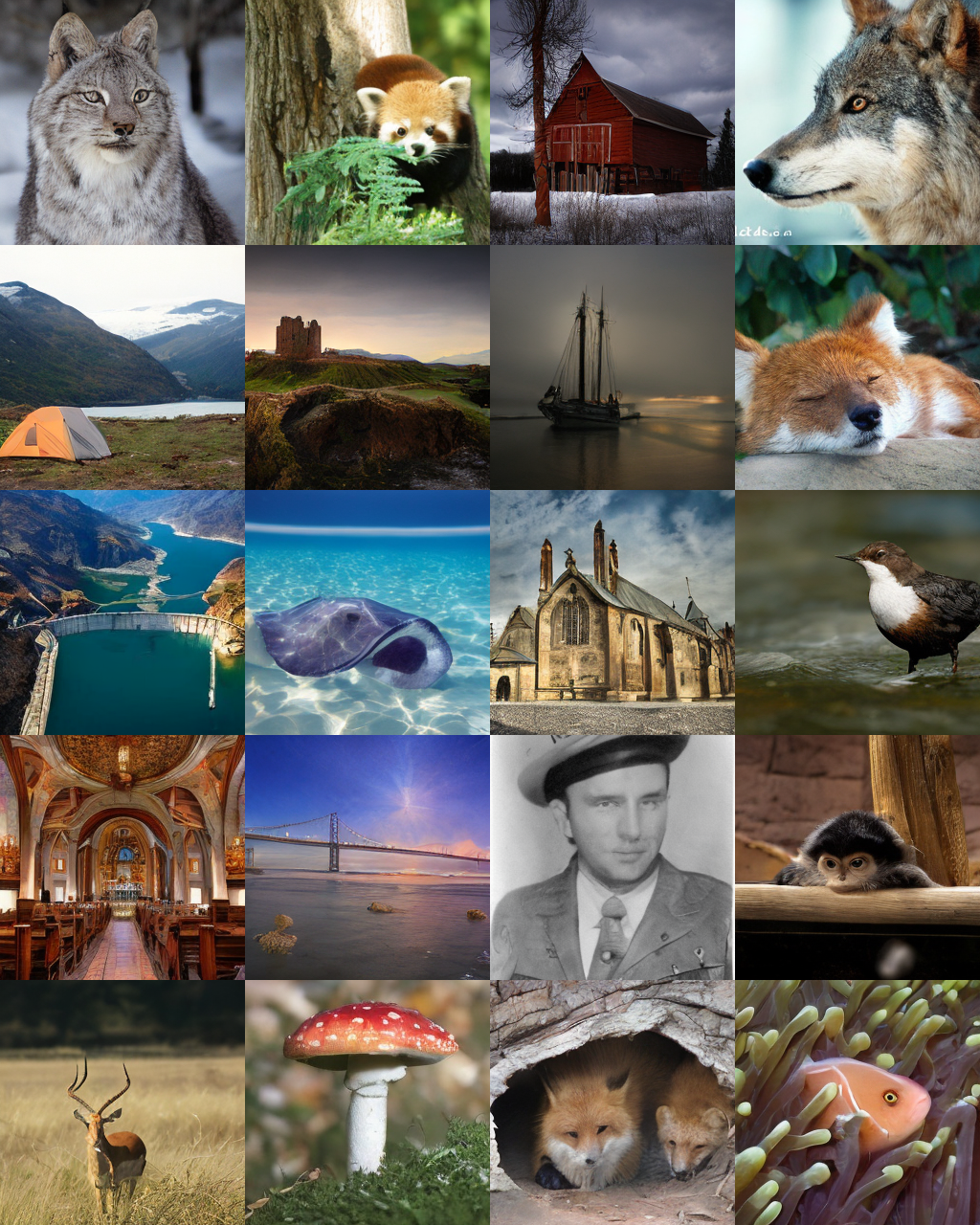}
    \caption{Additional generated samples on ImageNet $256\times 256$. The samples are from ViT backbone with 2.06 FID.}
    \label{fig:vit_in256}
\end{figure*}

We provide additional generated results in Figure~\ref{fig:CelebA64}-\ref{fig:vit_in256}.
Figure~\ref{fig:CelebA64} shows the generated samples with UNet backbone on CelebA $64\times 64$.
Figure~\ref{fig:vit_tn64} and Figure~\ref{fig:unet_in64} demonstrate the generated samples on conditional ImageNet $64\times 64$ benchmark with ViT-Large and UNet backbone respectively.
The visual results on CelebA $64\times 64$ and ImageNet $64\times 64$ are randomly synthesized without cherry-pick.

We also present some visual results on ImageNet $256\times 256$ with our model which can achieve the FID 2.06 in Figure~\ref{fig:vit_in256}.

\end{document}